\documentclass[final,11pt]{elsarticle}

\usepackage{graphicx}
\usepackage{amssymb}

\usepackage{subcaption}

\usepackage{url}

\journal{NAB Broadcast Engineering and Information Technology (BEIT) Conference}

\begin{document}

\begin{frontmatter}


\title{Implementing AI-powered Semantic Character Recognition in Motor Racing Sports}



\author{Jose David Fern\'andez Rodr\'iguez,\\
David Daniel Albarrac\'in Molina,\\
Jes\'us Hormigo Cebolla
}

\address{\textbf{Virtually Live}\\M\'alaga, Spain}

\begin{abstract}
Oftentimes TV producers of motor-racing programs overlay visual and textual media to provide on-screen context about drivers, such as a driver’s name, position or photo. Typically this is accomplished by a human producer who visually identifies the drivers on screen, manually toggling the contextual media associated to each one and coordinating with cameramen and other TV producers to keep the racer in the shot while the contextual media is on screen. This labor-intensive and highly dedicated process is mostly suited to static overlays and makes it difficult to overlay contextual information about many drivers at the same time in short shots.

This paper presents a system that largely automates these tasks and enables dynamic overlays using deep learning to track the drivers as they move on screen, without human intervention. This system is not merely theoretical, but an implementation has already been deployed during live races by a TV production company at Formula E races. We present the challenges faced during the implementation and discuss the implications. Additionally, we cover future applications and roadmap of this new technological development.
\end{abstract}


\end{frontmatter}


\section{Introduction}
\label{S:1}

In all domains of human endeavor related to technology, the history of the last decades has been one of relentless innovation, continually pushing the edges of what is possible. TV production is no exception to this trend and especially so in the use of special effects to aid and enhance sports storytelling.
 
The output of the production depends very much on the particular constraints of the program, and the biggest difference is between recorded and live production. For offline production, software tools like Mocha \cite{borisfx} enable producers to configure and add infographic elements that seamlessly integrate within the video feed, following the motion of drivers/players (or any other focus of attention in the corresponding sports) as their images move across the screen.
 
On the other hand, Live TV production has been much constrained in the style and configuration of infographic overlays for on-screen objects, which are typically limited to static, time-limited banners, such as lower thirds. These limitations stem from the practical impossibility of a human producer to recover target motions in real-time with a reasonable degree of accuracy and concurrently add infographic elements following such motions.
 
That said, there have been several sports where high-tech solutions have been applied to enable richer overlays, such as FoxTrax \cite{574652} or NASCAR’s RACEf/x \cite{milnes2001real}. FoxTrax was the technology behind the “glowing pucks” in Fox Sports broadcasts of NHL matches in the 1990s, while RACEf/x enables the overlay of special effects and infographic pointers onto NASCAR’s live feeds. However, these systems require very costly tracking equipment, a careful and slow setup on premises before each event (as the placement of the cameras has to be measured with a very high degree of accuracy and the racing tracks have to be mapped in 3D with very high precision) and are also very expensive to operate (in the range of hundreds of thousands of dollars per event), making them feasible in very few situations—basically only for massively popular sports repeatedly played in one or a few select places.

\begin{figure}[h]
\centering\includegraphics[width=0.5\linewidth]{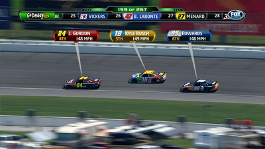}
\caption{Car Pointers overlaid onto the live feed with Nascar's Racef/x}
\end{figure}

In contrast, we present here a new approach for automatic and dynamic infographic overlays on live feeds based on deep learning, without any need for tracking equipment or previous setup on-site, and requiring only a modest computer system to operate.

\section{Deep Learning}

In the last few years, as computer hardware has become faster and cheaper, new machine-learning techniques such as deep learning has become feasible. Deep learning \cite{goodfellow2016deep} can be characterized as a relatively new term for a suite of methodologies based on artificial neural networks, which can automatically extract semantic information from various data sources (text, images, audio, video, sensor feeds) in ways that were previously not possible. While these methodologies can be traced back to the 1960s, it has been only relatively recently (from the 2010s) that they have become effective enough in relatively complex media (text, images, audio) to be easily adapted and deployed in business environments. This transformation has become possible because breakthroughs in computer technology (both in hardware and software) have enabled small computer systems (desktop-level and smaller) to perform the massive amounts of mathematical computations that are required to synthesize and use these deep learning systems at very high speeds.
 
Deep learning systems for object detection take an image and mark and label all objects they can recognize in the image, either as bounding boxes or as image masks. They do not perform open-ended object detection, but can detect only specific categories of objects. Object detection systems are crafted to detect some kinds of objects through a process known as training. During the training, the system is presented with a large corpus of images—the training set—and the semantic information about the objects of interest that appear in each image. The system gradually learns to correlate image content with its semantics in a way that eventually enables it to recover semantic information from images that it has not been trained on.

\begin{figure}[h]
\centering\includegraphics[width=0.5\linewidth]{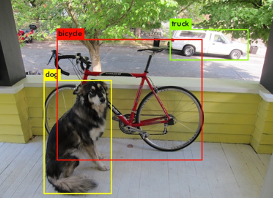}
\caption{Example of use of Yolo \cite{redmon2018yolov3}, a deep learning system for object detection. the system takes as input the image (without the boxes) and recognizes various objects in it, providing the type of object and placement for each instance. note that the system’s output is not this image; this is just a visualization of the result.}
\end{figure}

Currently, state-of-the-art deep learning systems for object detection can run on small form-factor computer boards at real-time frame rates (i.e., detecting objects on images at 25 images per second) for small image sizes, or on desktop computers at bigger image sizes (or at even faster frame rates). So, in principle, the technology is mature enough to be applied to live broadcasts.

\section{Deep Learning}

Using deep learning for object detection in a business environment presents challenges that are not usually found in the academic settings where such systems have been developed. Here, we will refer to car detection (in the context of a racing sport), but the same methodology can be applied to motorcycles, bikes, vessels or really any kind of racing vehicle.
 
If we want the deep learning system to detect the position and tell apart the cars driven by each pilot, we have to train it with a large corpus of images, with such cars appearing from a wide range of orientations and distances. The system can learn to tell apart the cars even if they all have the same shape, as long as they have different color patterns, even if these patterns are only slightly different. Of course, if the color patterns change significantly, the system may have to be retrained.
 
The training images have to be manually tagged, i.e., a human has to input the type and placement of all objects in each image. If the training images are not individual stills, but actual footage from sports events, this can be done in a very efficient way using already existing tracking software, such as Mocha.
 
To train the system, we need tagged images. These images can be sourced from existing footage, but if the color patterns of each object are occasionally changed, there might be very little footage, or none at all, before the next race. So, if we had to depend on just existing footage, we may not be able to detect objects for the first race, or for each race after a significant change in appearance.
 
However, this is not an unsolvable problem: Because deep learning systems can be tweaked to generalize quite a lot, it is possible to train them with synthetic images generated with 3D models of the cars and still be able to detect cars on real images. These synthetic images can be generated at a massive scale, with automatic object tagging (i.e., without any human in the loop). In fact, we have found that systems trained on a mix of real and synthetic images are able to detect cars in a more effective way than using only real or synthetic images.

\begin{figure}
    \begin{subfigure}{.49\textwidth}
        \centering
        \includegraphics[height=4cm, width=.9\linewidth]{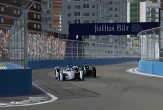}
        \caption{}
        \label{fig:drawinga}
    \end{subfigure}
    \begin{subfigure}{.49\textwidth}
        \centering
        \includegraphics[height=4cm, width=.9\linewidth]{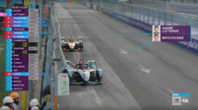}
        \caption{}
        \label{fig:drawingb}
    \end{subfigure}
    \caption{Drawing (a) shows representation of angular components, based on extrinsic Euler angles, to describe the object's cuboid hull to be inferred by the neural network. Drawing (b) represents the regions assigned to the the 6 nearest priors out of the 18. The 2 front-facing regions (blue and green) are horizontally sectioned for clarity.}
    \label{fig:angles}
\end{figure}

Typically, we have observed that for the purposes of car detection, we can train our systems in one or two days. While this requires a measurable lead time before the system is ready to detect objects, it is short enough that this has not presented an operational problem.

\section{Workflow}

In practical terms, the system we propose works as follows:

\begin{itemize}
    \item The live video footage is fed to a computer (the back-end) with a video capturing card. In this computer, the deep learning system is applied to every incoming image frame. The object detections are sent to the front-end. The back-end requires either a big and powerful general-purpose computer or a small and specialized one (such as a Nvidia Jetson Xavier board \cite{ditty2018nvidia}).
    \item The front-end offers a powerful user interface to configure the appearance of the infographic elements to be displayed for each detected participant. It receives data about the object detections from the front-end and renders the corresponding infographic elements for each object. This is the part of the system intended to be operated by TV producers. It can either take the video footage feed and render the infographic elements directly onto it or output just the infographic elements, to merge them with the footage at a later stage of the live production pipeline.
\end{itemize}

The system is intended so that the live TV producer uses the front-end in real time to tweak infographic configurations and generally operate the whole system. If the TV producer can use a desktop computer to run the front-end, the back-end can be run on the same computer (even a compact one).
 
However, if operational constraints make it more convenient to use a lightweight laptop for the front-end, the back-end might not be able to run on it (it requires a powerful GPU card). In that case, it can be hosted on another computer, even a small form-factor one, and it can communicate with the front-end using a computer network connection.

\begin{figure}[!htb]
\minipage{0.32\textwidth}

        \centering
        \includegraphics[height=2.5cm]{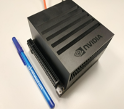}
        \caption{NVidia Jetson xavier is a small form-factor computer able to run the back-end}

\endminipage\hfill  
\minipage{0.32\textwidth}
    
        \centering
        \includegraphics[height=2.5cm]{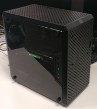}
        \caption{a compact desktop computer (36x37x22cm) able to run both the back-end and the front-end}

\endminipage\hfill      
\minipage{0.32\textwidth}    
    
        \centering
        \includegraphics[height=2.5cm]{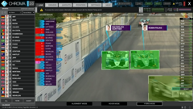}
        \caption{a screenshot of the front-end.}
    
\endminipage\hfill      

\end{figure}

\section{Future Enhancements}
Currently, the system can detect objects in each image, detect each object’s class (in our case, the driver associated with the car’s livery) and point to the object’s center. We are currently working on two kinds of enhancements: enhanced frame-to-frame tracking and automatic inference of 3D parameters:

\begin{itemize}
    \item As the detection is done separately for each video frame, the deep learning system may sometimes lose track of a driver for a few frames. While this can be solved with some simple heuristics, these introduce a level of complexity that has to be exposed in the front-end, making the operator’s task more complex that it should be. We are working on enhancing the deep learning system so it can reason with several frames’ worth of information and can keep track of drivers across frames without the need for additional heuristics.
    \item While the detection system is effective enough, the range of effects that it enables is rather limited. The reason is that, while the system can detect objects, it knows nothing about the object’s 3D orientation and position concerning the camera. Being able to infer these 3D parameters would enable new and exciting possibilities, such as infographic elements pointing to specific parts of the car (the tires, the driver) or special effects being applied to specific parts of the car (such as adding special effects in the car’s tires or the car’s rear). These kinds of special effects can be routinely added as a post-production step, but the proposed enhancement to our deep learning system would enable such effects to be applied in real time. 
\end{itemize}

Additionally, we are also working on optimizing the back-end software to be able to infer with larger images to detect smaller objects, still at real-time frame rates.

\section{Conclusion}

It has been only in recent years that deep learning has started to make breakthroughs in various difficult problems such as object detection, so it is only now that corporations have started to apply deep learning to solve real business needs. In our specific case, we have reduced the cost to deploy and operate state-of-the-art, real-time detection systems by orders of magnitude. Additionally, as our system does not require drivers to carry sophisticated and bulky location hardware or be in a tightly mapped environment, it can be applied to any racing sport. Finally, as we develop and enhance the capabilities of our system, we will be able to automatically add new kinds of infographic elements and special effects in live video feeds, previously only possible in post-production.

\begin{figure}
    \begin{subfigure}{.49\textwidth}
        \centering
        \includegraphics[height=4cm, width=.9\linewidth]{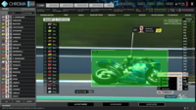}
        \caption{}
        \label{fig:drawinga2}
    \end{subfigure}
    \begin{subfigure}{.49\textwidth}
        \centering
        \includegraphics[height=4cm, width=.9\linewidth]{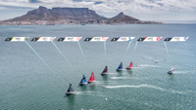}
        \caption{}
        \label{fig:drawingb2}
    \end{subfigure}
    \caption{Screenshot (a) of the front-end showing results of a back-end trained to detect MotoGP drivers. (b): mock-up showing a possible final rendering for The Ocean Race.}
    \label{fig:angles2}
\end{figure}

\pagebreak




\bibliographystyle{elsarticle-num-names}
\bibliography{bibliography.bib}

\begin{thebibliography}{6}
\expandafter\ifx\csname natexlab\endcsname\relax\def\natexlab#1{#1}\fi
\providecommand{\url}[1]{\texttt{#1}}
\providecommand{\href}[2]{#2}
\providecommand{\path}[1]{#1}
\providecommand{\DOIprefix}{doi:}
\providecommand{\ArXivprefix}{arXiv:}
\providecommand{\URLprefix}{URL: }
\providecommand{\Pubmedprefix}{pmid:}
\providecommand{\doi}[1]{\href{http://dx.doi.org/#1}{\path{#1}}}
\providecommand{\Pubmed}[1]{\href{pmid:#1}{\path{#1}}}
\providecommand{\bibinfo}[2]{#2}
\ifx\xfnm\relax \def\xfnm[#1]{\unskip,\space#1}\fi
\bibitem[{BorisFX(2019)}]{borisfx}
\bibinfo{author}{BorisFX}, \bibinfo{title}{Mocha, a planar tracking tool},
  \bibinfo{year}{2019}. \URLprefix
  \url{https://borisfx.com/products/mocha-pro/}.
\bibitem[{{Cavallaro}(1997)}]{574652}
\bibinfo{author}{R.~{Cavallaro}},
\newblock \bibinfo{title}{The foxtrax hockey puck tracking system},
\newblock \bibinfo{journal}{IEEE Computer Graphics and Applications}
  \bibinfo{volume}{17} (\bibinfo{year}{1997}) \bibinfo{pages}{6--12}.
\bibitem[{Milnes and Ford(2001)}]{milnes2001real}
\bibinfo{author}{K.~Milnes}, \bibinfo{author}{T.~Ford},
\newblock \bibinfo{title}{Real-time gps fx},
\newblock \bibinfo{journal}{GPS World} \bibinfo{volume}{12}
  (\bibinfo{year}{2001}) \bibinfo{pages}{12--26}.
\bibitem[{Goodfellow et~al.(2016)Goodfellow, Bengio, and
  Courville}]{goodfellow2016deep}
\bibinfo{author}{I.~Goodfellow}, \bibinfo{author}{Y.~Bengio},
  \bibinfo{author}{A.~Courville}, \bibinfo{title}{Deep learning},
  \bibinfo{publisher}{MIT press}, \bibinfo{year}{2016}.
\bibitem[{Redmon and Farhadi(2018)}]{redmon2018yolov3}
\bibinfo{author}{J.~Redmon}, \bibinfo{author}{A.~Farhadi},
\newblock \bibinfo{title}{Yolov3: An incremental improvement},
\newblock \bibinfo{journal}{arXiv preprint arXiv:1804.02767}
  (\bibinfo{year}{2018}).
\bibitem[{Ditty et~al.(2018)Ditty, Karandikar, and Reed}]{ditty2018nvidia}
\bibinfo{author}{M.~Ditty}, \bibinfo{author}{A.~Karandikar},
  \bibinfo{author}{D.~Reed},
\newblock \bibinfo{title}{Nvidia’s xavier soc},
\newblock in: \bibinfo{booktitle}{Hot Chips: A Symposium on High Performance
  Chips}, \bibinfo{year}{2018}.

\end{thebibliography}







\end{document}